\documentclass[letterpaper, mathptm,psfig,10 pt, conference]{ieeeconf}  
\IEEEoverridecommandlockouts                              
\usepackage{graphics} % for pdf, bitmapped graphics files
\usepackage{graphicx}
\usepackage{subcaption}
\usepackage[font=footnotesize]{caption}
\usepackage{amsmath} % assumes amsmath package installed
\usepackage{amssymb}  % assumes amsmath package installed
\usepackage{bbm}
\usepackage{bbold}
\usepackage{algorithm, algorithmic}
\usepackage{dsfont}

\usepackage{cite}
\title{\LARGE \bf
Top-K Ranking Deep Contextual Bandits for Information Selection Systems
}

\author{Jade Freeman$^{1}$ and Michael Rawson$^{2}$
\thanks{$^{1}$Jade Freeman is with DEVCOM Army Research Laboratory, 2800 Powder Mill Road, Adelphi, MD 20883, USA
        {\tt\small jade.l.freeman2.civ@army.mil}}%
\thanks{$^{2}$Michael Rawson is with the Department of Mathematics, University of Maryland,
        College Park, MD  20742, USA
        {\tt\small rawson@math.umd.edu}}
}

\begin{document}

\maketitle
\thispagestyle{empty}
\pagestyle{empty}

%%%%%%%%%%%%%%%%%%%%%%%%%%%%%%%%%%%%%%%%%%%%%%%%%%%%%%%%%%%%%%%%%%%%%%%%%%%%%%%%
\begin{abstract}
In today's technology environment, information is abundant, dynamic, and heterogeneous in nature. Automated filtering and prioritization of information is based on the distinction between whether the information adds substantial value toward one's goal or not. Contextual multi-armed bandit has been widely used for learning to filter contents and prioritize according to user interest or relevance. Learn-to-Rank technique optimizes the relevance ranking on items, allowing the contents to be selected accordingly. We propose a novel approach to top-K rankings under the contextual multi-armed bandit framework. We model the stochastic reward function with a neural network to allow non-linear approximation to learn the relationship between rewards and contexts.  We demonstrate the approach and evaluate the the performance of learning from the experiments using real world data sets in simulated scenarios. Empirical results show that this approach performs well under the complexity of a reward structure and high dimensional contextual features. 

\end{abstract}

\bigskip
%%%%%%%%%%%%%%%%%%%%%%%%%%%%%%%%%%%%%%%%%%%%%%%%%%%%%%%%%%%%%%%%%%%%%%%%%%%%%%%%
\section{INTRODUCTION}

In the current era of perpetual influx of information, decision makers and information analysts are faced with challenges under fixed time and resources. Consider a scenario where the environment is isolated with low network resources and information-driven decisions must be made in a timely manner. Even with a finite number of information sources, computational process, network capacity, and human cognitive abilities can be limited in time to download, evaluate, and make judgments on available information in full. The best selection policy is to filter out the information that is of little use or interest to the user. To this end, a variety of expert systems has been developed in recent decades as a mean for coping with cognitive and systematic information overload problems.

 Many information selection systems are built on the concept of an expert system. Under shifting conditions and priorities in the environment, the user requirement for information can vary from one situation to another. An efficient and effective expert system learns and adapts to each situation and identifies and provides the most useful and relevant information. An expert system based on reinforcement learning applies previous responses to already seen, related  or similar objects. The user interest is inferred implicitly via user interaction patterns on the content and historically observed utility of the information. The contextual information, which represents a set of observable factors about the users, the environment and the objects, can impact the  relevancy, preference, and decision on the objects. The context can be represented categorically (e.g., class, occupation) or numerically (e.g., temperature, distance). The context can come from unstructured contents such as messages or images. 

Determining a single best object is simpler than multiple objects.    When the system needs to provide a set of  most relevant or user-preferred information objects, the potential candidates of the information becomes a power set. As the problem of choosing the optimum set of $K$ objects from a set $D$ is equivalent to the maximum coverage problem, evaluating each collection of power sets $A_1, A_2, \dots \in \mathcal{P}(D)$ of $K$ elements is combinatorial and becomes NP-hard.  

In this work, we present a novel approach for implementing a deep learning to extend the contextual multi-armed bandit decision strategies to optimize the reward approximation based on the contexts from the contents with high dimensional features and Learning-to-Rank (LTR) technique to select the optimal set on information objects. We compare our model against other baseline approaches to show effectiveness in simulations.

The remainder of this paper is organized as follows. Section II provides an overview of multi-arm bandit method and some of the selection strategies. Section III describes some of the existing methods relevant to this work. Section IV describes overall approach to selecting multiple objects as ranking in contextual multi-arm bandit framework and provides the implementation of deep learning. Section V provides the demonstration and discusses the results from the experiments. We then conclude with final remarks on future work in Section VI.

\section{PRELIMINARY ON MULTI-ARMED BANDIT}
The Multi-Armed Bandit (MAB) is a decision problem where the decision maker is
presented with multiple arms, from which he or she is to choose an arm\cite{Gittins1979BanditPA}, \cite{Bather1990MultiArmedBA}. In applications, the ``arms" represent a set of objects with each having an unknown distribution of reward upon selection \cite{Qin2014ContextualCB}. MAB framework is applied in reinforcement learning to dynamically match an object to the user's interest from the previously seen behavior. A reward function is used to measure the quality (relevance) of the selected object based on the observed user feedback. The goal of the learning algorithm is to maximize the rewards or equivalently to minimize regrets over time.

\begin{figure}[b]
  \centering{
  \includegraphics[width=0.9 \linewidth,clip]{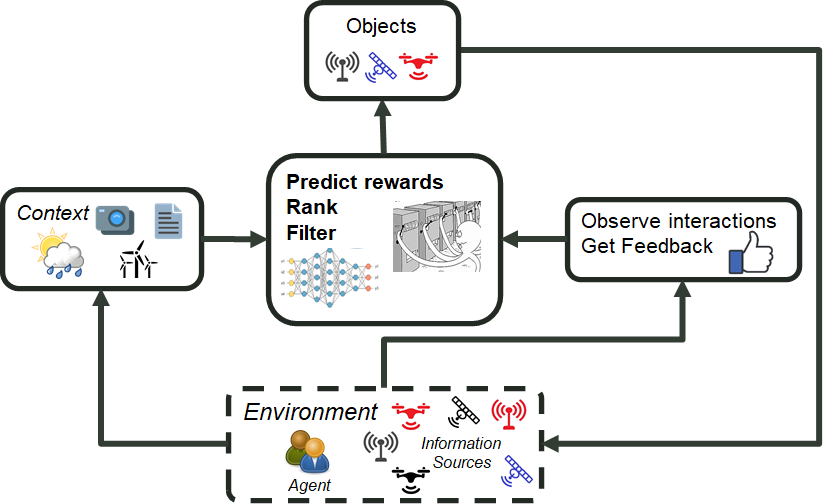}}
  \caption{General framework of top-K deep contextual bandit learning}
  \label{mab_voi}
\end{figure}

The bandit algorithm employs the selection strategy that balances the decisions between exploration of the unseen and exploitation of previously successful arms. One of the most popular exploration-exploitation strategy \cite{Langford2007TheEA} is $\epsilon$-greedy method where the exploration is uniform over arms with $\epsilon > 0$ probability. For each round $t=1,2...$,  a random probability from an uniform distribution is observed. If the probability is greater than $1-\epsilon$, then the best arm with the highest average reward is chosen, or otherwise, an arm is randomly chosen. $\epsilon-greedy$ does not converge to optimum and by making the $\epsilon$ dependent on time, decaying $\epsilon-greedy$ emphasizes exploitation over time as the learning becomes confident.  

Another well-known exploration-exploitation strategy is based on Bayesian Thompson sampling \cite{Thompson1933ONTL}, \cite{Agrawal2012AnalysisOT}. Let $\mu_i$ denote the expected reward of $r_i$ and observe reward $\hat{r}_t$ from the selected arm  $a_i, i=1,...,N$, and  $\mu_i \sim  p(\theta_i)$, some prior distribution with a parameter $\theta_i$. Let $a_i^* = \arg\max_i \mu_i$  denote the arm with the highest expected mean given parameter $\mathbf{\theta}_i$. Let $F_t=p(\mu | \theta,r_i)$ denote the current posterior density over $\mathbf{\theta}$. At each round $t$, select the arm $a_i$ having largest expected mean given this posterior, i.e., $ P(a_i^*|\theta(t))$, $\theta(t) \sim F_t$. The posterior distribution is updated with $\hat{r}_t$ and the process in continued iteratively.

\section{RELATED WORK}
Contextual multi-armed bandit, or contextual bandit in short, uses MAB strategy for the estimation and learning the rewards based on feedback, while adapting to observed contexts about the information and the environment \cite{Auer2002UsingCB}. Contextual bandit allows the learning process to incorporate features and context regarding the user, the objects, and the decision making environment, further enhancing the learning on which object to deliver to the user. In contextual bandit, assuming a linear relationship between the context and the reward is often sufficient and simple and straightforward in implementation. However, it can result in unsatisfactory performance when linearity assumption is invalid. The deep neural network (DNN) has been successful in approximating non-linear value functions, proven to be a powerful function approximator, and one can learn the associations between the complex states to estimates of expected return \cite{Riquelme2018DeepBB}. Further, DNN allows generalization of  the states without domain specific knowledge and learn rich domain representations from raw, high-dimensional inputs \cite{Mnih2015HumanlevelCT}, \cite{Silver2016MasteringTG}, \cite{Zahavy2019DeepNL}. Deep bandit learning has been studied in Bayesian framework, which utilizes uncertainty on the model in exploration \cite{Guo2020DeepBB}. Collier and Llorens \cite{Collier2018DeepCM} proposed a contextual multi-armed bandit method in a Bayesian neural network framework with dropout method. 

The approach for selecting a set of arms in MAB framework was proposed by Chen et al. \cite{Chen2013CombinatorialMB} via approximation-oracle method. Their $(\alpha, \beta)$-approximation oracle uses the means of the distributions of arms and outputs a super arm that with probability $\beta$ generates an $\alpha$ fraction of the optimal expected reward and the algorithm is to minimize the approximated regret.  Learning-to-Rank (LTR) technique \cite{10.1145/2507157.2508063} is a ranking-based information retrieval technique for selecting multiple items from upon user query or user preference. Hu et. al \cite{Hu2018ReinforcementLT} proposed deterministic policy gradient method to learn optimal ranking policy in reinforcement learning which maximizes the expected cumulative rewards in each time step. Intayoad et al. \cite{Intayoad2020ReinforcementLB} integrated contextual information into the value function in reinforcement learning and utilizes correlation matrix on previous actions to determine top-K rank actions to select the most optimal action values while implementing $\epsilon$-greedy method to ensure the exploration of all states and actions.  

\section{DEEP CONTEXTUAL BANDIT FOR TOP-K RANKING}
 We approach the problem of learning to correctly predict information objects from a finite set, which a user would choose in a stationary state, as a contextual multi-armed bandit (CMAB) for ranking given historical interactions and the context. To formalize, suppose there are $n$ number of objects in $D$. The objects must be prioritized and filtered down to  a subset with $K$ objects.

Consider the ranking of the objects in $D=\{d_1,...,d_n\}$ at each round $t$, a subset $A_t\subset D$ and $A_t=\{d_{(1),t},...,d_{(k),t}\}$, where $d_{(i),t}$ is the $i_{th}$ ranked  object at time t. At time $t$, the context $X_{i,t}$ is observed and $R(d_{i,t} \mid X_{i,t})$ is the reward of $d_{i,t}$. Denote the expected reward from $d_{i,t}$ as $E\left[R(d_{i,t}) \mid X_{i, t} \right]=\mu_{d_{i,t}}$. 

Using the context features and the estimated rewards, the $K$ objects with highest rewards are predicted by running a MAB algorithm. Let $d_t^\ast=argmax_{d_t}  \left[ \mu_d\right] $ be the selections that maximizes the total reward from MAB exploitation-exploration schemes at time $t$. We maximize the expected cumulative rewards $\sum_{t=1}^{T} \sum_{i=1}^n E\left[R(d_{i,t} \mid X_{i, t})\right]$. This strategy is equivalently reformulated to minimize the potential regret $\mu_ {d_t^*}-\mu_{d_t}$ by  choosing the optimal set so that the cumulative regrets $Reg(T)=\sum_t(\mu_ {d_t^*}-\mu_{d_t})$ is as small as possible over time $T$. $Reg(T)$ is also used as a metric to quantify the performance of learning up to the time step T. At each round $t$, $d_{(i),t}^\ast$ is determined from $D$ by successively excluding previously selected $d_{(j),t}^\ast$, $j=1,...,i-1$. With $A_t^\ast$,  DNN is trained to approximate the reward function based on the contexts on $A_t^\ast$ and their observed rewards. See Algorithm 1 for the iterative process.
\begin{algorithm}[H]
\smallskip
\begin{algorithmic} 
\STATE $D=\{d_1,...,d_n\}$ 
\STATE $K$=Number of arms (objects) to choose from $D$
\STATE $X_{t,i}$ is the $i^{th}$ context vector at time step $t$   
\STATE NN = Neural Network   
\STATE Reward : $D \rightarrow \mathds{R}$ 
\FOR{$t = 1 ... T$} 
 \FOR{$i = 1 ... K$} \STATE $\hat{\mathds{E}}(arm_i) = NN(X_{t,i}) $\ENDFOR 
 \FOR{$i = 1 ... K$} \STATE $\tilde d_{t,i}$ = Generated from MAB algorithm with $\hat{\mathds{E}}(arm_i)$ on $D\setminus \tilde d_{t,1:i-1} $ \ENDFOR 
 \FOR{$i = 1 ... K$} \STATE $R_{t,i} = Reward(\tilde d_{t,i})$ \ENDFOR \\
Train NN($NN, input=X_{1:t,\tilde d_{t,:}}, output=R_{1:t,:} )$  
\ENDFOR  
\end{algorithmic}
 \caption{Deep Neural Net for Top-K Selection}
\end{algorithm}
\section{Experiments and Results}
We demonstrate the implementation of our approach on two tasks with real-world data sets. We set up the classification data sets in a reinforcement learning process. We evaluate the learning performances on $\epsilon-greedy$ with a positive and decaying $\epsilon$ probabilities and Thompson's sampling algorithms, varying degrees of complexity of reward structures and the size of contexts. We run the experiments on the random selection of arms as a baseline for the MAB algorithms. Also, linear regression and neural-linear methods provided the baselines for DNN implementation. These two methods do not explore and select purely greedy. Under the neural-linear, the feature representations are learned from a neural network which is applied to the linear regression model on top of the last hidden layer of a neural network. It was first proposed for use with Bayesian optimization \cite{Snoek2015ScalableBO} and has been applied in bandit problems \cite{Riquelme2018DeepBB}, \cite{Zahavy2019DeepNL}. The neural networks are trained with Adam\cite{Kingma2015AdamAM} optimizer over 16 epochs minimizing mean square error.  We vary the bandit parameters and neural net tuning-parameters including number of arms, $K$, and number of neurons in the network layers. Finally, to represent real-world scenario, we add a scaled noise term to the reward functions. 
\begin{figure}
%\smallskip
\center
\subcaptionbox{Snapshot of mushroom classification data}[.8\linewidth][l]{%
  \includegraphics[width=1\linewidth,clip]{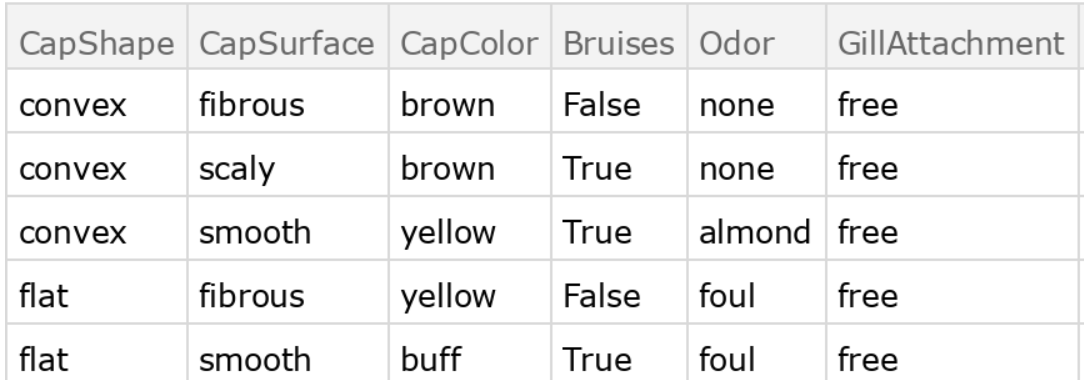}}
\bigskip

\subcaptionbox{Samples from MNIST digits images}[.8 \linewidth][l]{%
 \includegraphics[width=1\linewidth,clip]{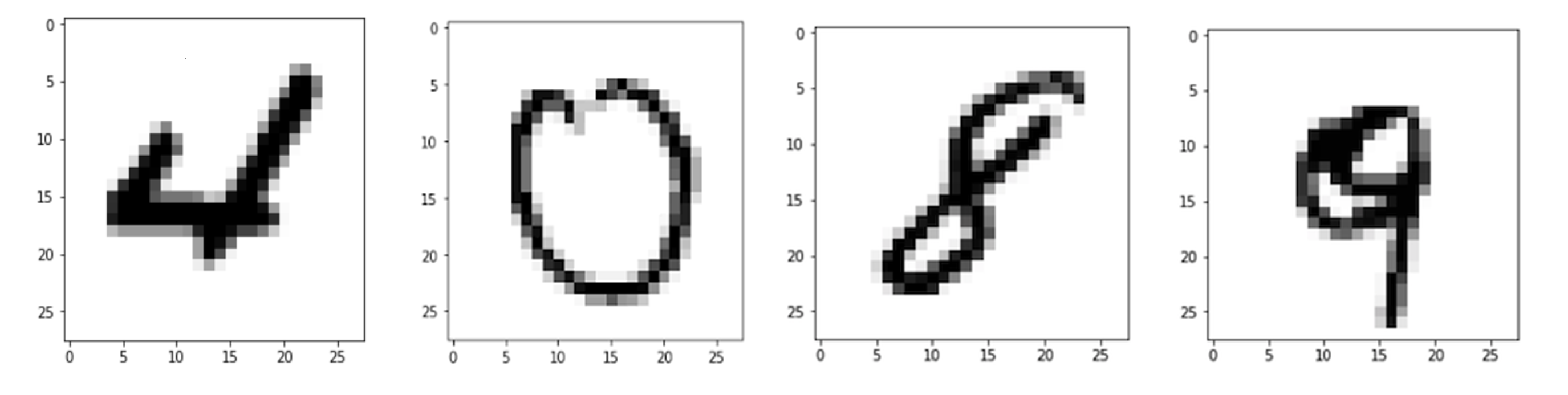}}
 \caption{Examples of data used in experimentation}
 \label{data}
  \end{figure}
\subsection{Learning to Choose Edible Mushrooms}
In this experiment, we use the mushroom data set (Fig. 2(a)) from the University of California-Irvine (UCI) Machine Learning Repository\cite{Dua:2019}. The goal is to choose $K$-mushrooms that are most likely edible from a collection of $n$ mushrooms based on their features and past experiences. The data set includes the labels on whether the mushroom is edible or poisonous with 22 characteristic features for each mushroom. The reward is 1 if the selected mushroom is edible and 0 if poisonous.  The reward noise term $\eta\sim Normal(0,1)$  is scaled by 1/2 and thus, the cumulative rewards $R(A_t) = \sum_{i \in A_t} [\frac{\eta}{2} + \mathds{1}_{X_{t,i}\ edible}]$. We maximize the cumulative rewards from selected K mushrooms given the context from the mushroom features. In simulating mushroom collections, we balance the data set so that the chance of a context vector (mushroom) being edible is $K$/$n$. The neural network has 2 dense layer size either 100 or 1,000 with the ReLU activation function (ReLU), followed by single node output. 

Fig. 3 shows the plots of cumulative regrets and rewards over time on selecting 3 edible mushrooms ($K$=3) out of 30 or 50 mushrooms (n=30, 50) and the number of neurons for the neural network model was 100. It took more time steps to learn to select the edible mushrooms with the larger group of mushrooms at 50 compared to 30, which follows the intuition of the need to explore more mushrooms at a given time. In this experiment, the linear and neural-linear model performed better than bandit methods on DNN.

\begin{figure*}[t]
\centering
\subcaptionbox{Cumulative regrets. Number of arms n=30}[.4\linewidth][c]{%
  \includegraphics[width=1.1\linewidth,clip]{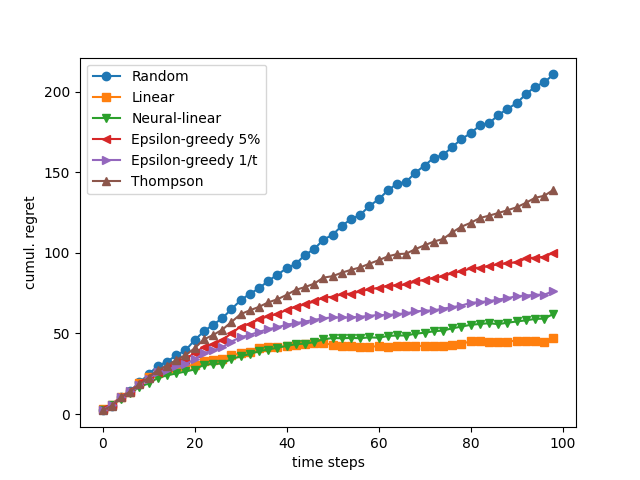}}
\quad
 \subcaptionbox{Cumulative rewards. Number of arms n=30}[.4\linewidth][c]{%
  \includegraphics[width=1.1\linewidth,clip]{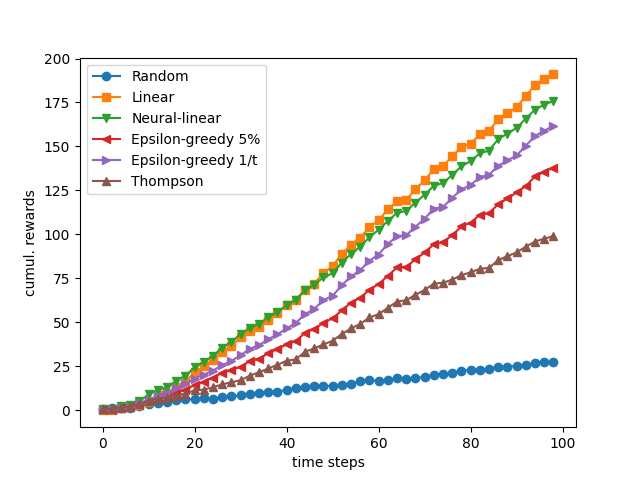}}

  \subcaptionbox{Cumulative regrets. Number of arms n=50}[.4\linewidth][c]{%
  \includegraphics[width=1.1\linewidth,clip]{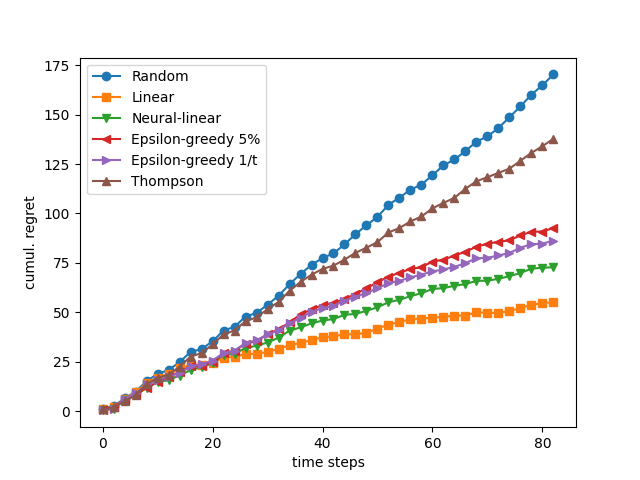}}
  \quad
    \subcaptionbox{Cumulative rewards. Number of arms n=50}[.4\linewidth][c]{%
  \includegraphics[width=1.1\linewidth,clip]{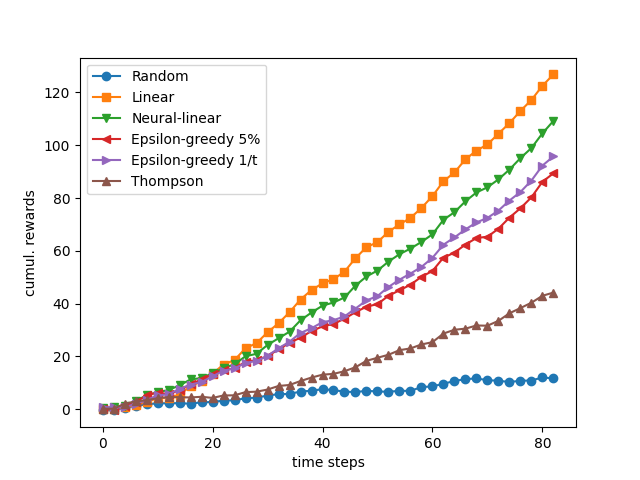}}
  \label{mushrooms100}
%   \skip
   \caption{Plots of edible mushroom selections. Choose top K=3. Number of neurons=100.}
 \end{figure*}

 \begin{figure*}
 \centering
\subcaptionbox{Cumulative regrets. Number of arms n=30}[.4\linewidth][c]{%
  \includegraphics[width=1.1\linewidth,clip]{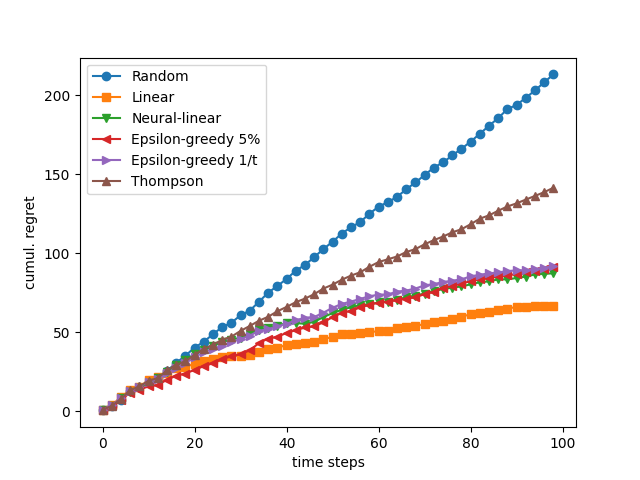}}
  \quad
  \subcaptionbox{Cumulative rewards. Number of arms n=30}[.4\linewidth][c]{%
  \includegraphics[width=1.1\linewidth,clip]{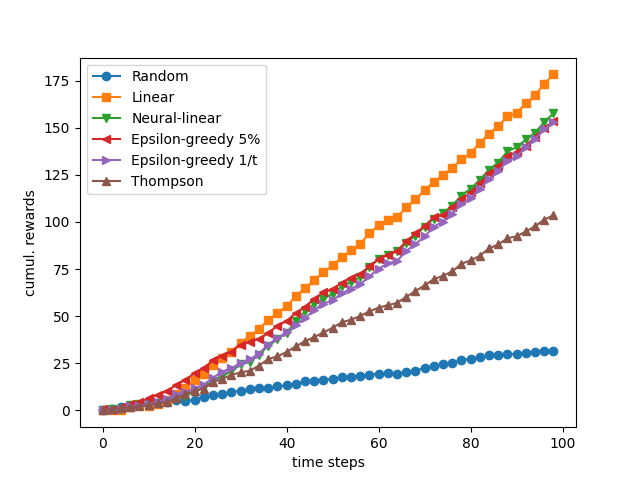}}
  
\subcaptionbox{Cumulative regrets. Number of arms n=50}[.4\linewidth][c]{%
  \includegraphics[width=1.1\linewidth,clip]{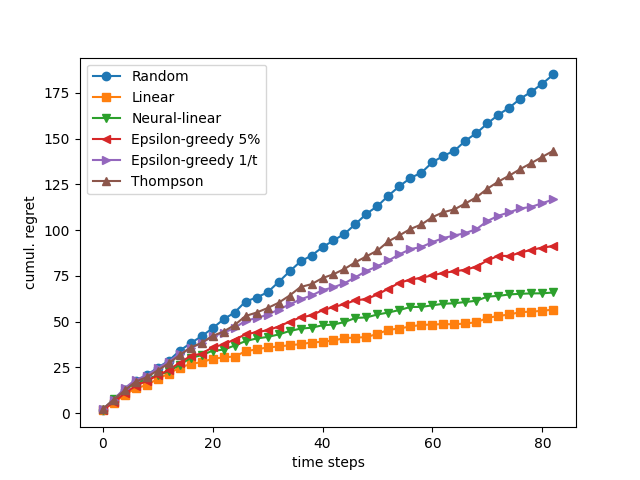}}
  \quad
  \subcaptionbox{Cumulative rewards. Number of arms n=50}[.4\linewidth][c]{%
  \includegraphics[width=1.1\linewidth,clip]{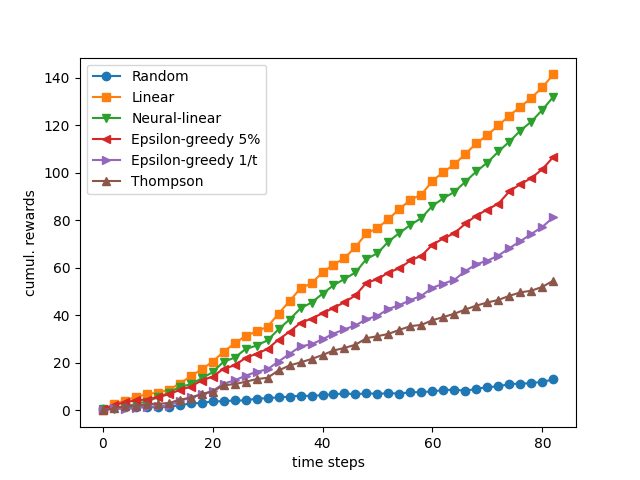}}
   \label{mushrooms1000}
%   \skip
    \caption{Plots on learning edible mushrooms. Choose top K=3. Number of neurons=1,000.}
  \end{figure*}
 \captionsetup[subfigure]{labelformat=simple}
\renewcommand*{\thesubfigure}{(\alph{subfigure})}
Fig. 4 shows the plots of cumulative regrets and rewards over time on selecting edible 3 mushrooms out of 30 or 50 mushrooms, but the number of neurons for the neural network model was 1,000. However, the increased neurons did not improve the results. With 30 mushrooms, $\epsilon-greedy$ algorithms performed nearly as well as the neural-linear. However, this performance did not hold for 50 mushrooms. 
 
Overall, our approach was able to learn to select correct mushrooms better than random guess. The linear model-based algorithms performed well consistently, with a linear regression being the best and then the neural-linear the next best.  Thompson sampling method performed the worst except for the random guesses. The increase in the neurons in the neural network  did not appear to improve the performance.
 
\begin{figure*}[t]
\centering

\subcaptionbox{Cumulative regrets. Number of neurons=100}[.4\linewidth][l]{%
  \includegraphics[width=1.1\linewidth,clip]{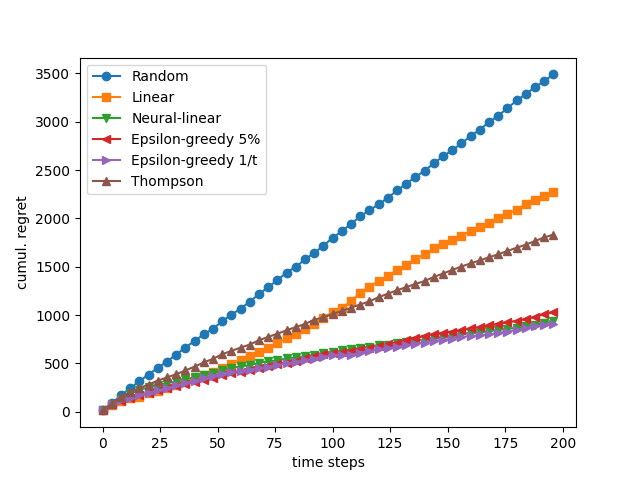}}
\quad
\subcaptionbox{Cumulative rewards. Number of neurons=100}[.4\linewidth][l]{%
  \includegraphics[width=1.1\linewidth,clip]{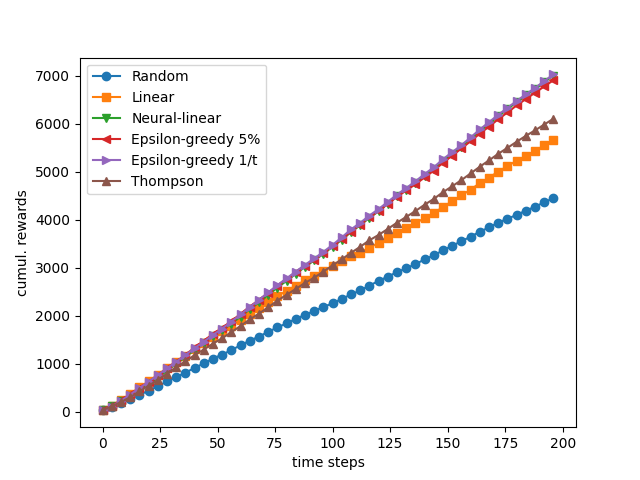}}
  
  \subcaptionbox{Cumulative regrets. Number of neurons=1,000}[.4\linewidth][l]{%
  \includegraphics[width=1.1\linewidth,clip]{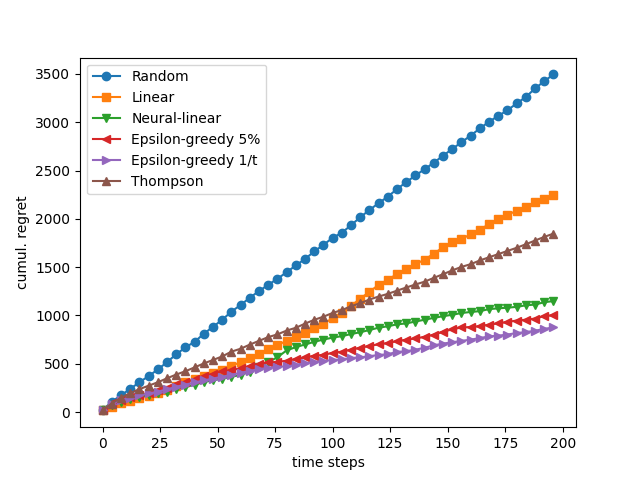}}
\quad
\subcaptionbox{Cumulative rewards. Number of neurons=1,000}[.4\linewidth][l]{%
  \includegraphics[width=1.1\linewidth,clip]{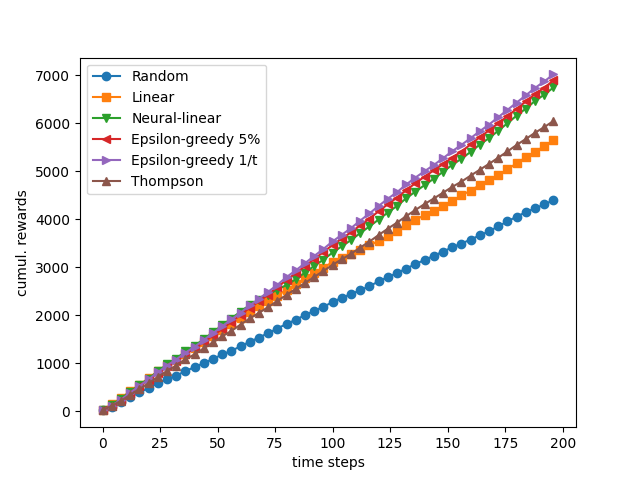}}
   
%   \skip
  \caption{Plots on largest values from MNIST images. Number of arms n=20. Choose top K = 5.}
  \label{MNIST}
  \end{figure*}

\subsection{Learning to Choose Top-k Largest Values from MNIST Images}
In this experiment, we setup a bandit problem using the digit samples from the MNIST data sets (Fig. 2(b)) \cite{deng2012mnist}. Let $n$ be the number of arms which each gets a context image and let $K$ be the number of selected images.
% , and let $d_i$ be an image object of an integer between 0 and 9. 
The goals is identify the top-K largest integers given $n$ images. The image features are used as the contexts X and the reward function $R$ on the set $A$ at t is defined as $R(A_{t})= \sum_{i \in A_t} [ I(X_{t,i}) +p\cdot\eta_i]$ where $I(\cdot)$ is the digit of the image
% from the $d$ inferred from its feature vector X 
and,  $\eta$ is the $Normal(0,1)$ noise term in the reward, scaled by $p$=2. The reward is maximized by correctly learning the largest values from the set of images of digits in $D$. We balance the data set so that each context vector is grey-scale, normalized between 0 and 1, and has uniform probability among digits. In this experiment, the number of images is $n$=20 and the number of arms chosen $K$ = 5. The deep learning is implemented with a convolutional neural network architecture which has 3 layers of convolution, max-pooling, and ReLU, then followed by a dense layer of size 100 or 1,000 with ReLU, and single node output. 

Fig. 5 shows the plots on the cumulative regrets and rewards. Over time, the linear model begins to under-perform compared to the neural network models, neural-linear, $\epsilon-greedy$, and Thompson sampling algorithms. Thompson sampling method did not show a better performance compared to neural-linear and $\epsilon-greedy$. In particular, $\epsilon-greedy$ at 5 percent and decaying rate over time showed the best performances. The neural-linear performed better than the linear method. The results from this experiment warrants further investigation by examining the behavior on longer training steps and varying number of neurons.
\subsection{Discussion}
The two experiments presented the results from contrasting example data sets on top-K ranking deep contextual bandit methods. The mushroom data experiment has the context that is readily interpretable, has small number of features, and a simple reward structure (the sum of weighted binary values). Although there was no a-priori reason to believe that relationship between the reward and the contexts for each mushroom is linear, the linear method performed better than DNN. It follows that the size of neurons in the model did not make much impact on the performances and the implementation of deep learning rather degrades the performance.

On the other hand, MNIST data experiment has the context with high dimensional features as an input and thus, the reward of selecting optimal $K$ integers is a non-linear. The results showed that the linear model-based algorithms did not perform well compared to the DNN-based algorithms. Surprisingly, simpler $\epsilon-greedy$ algorithms outperformed more sophisticated Bayesian Thompson sampling approach on the neural network. Neural-linear has shown to perform well despite it showed a drop in performance compared to $\epsilon-greedy$.  Since neural-linear does not implement any bandit exploration scheme, the learning was limited to previously observed optimum sets. The learning performance could depend on the choices of neural network model tuning parameters, and it needs further investigation on how these parameters can affect the learning performance. 

\section{CONCLUSION AND FUTURE WORK}
In this work, we showed that our approach learns the $K$-most user-relevant of information objects from a given finite set as a top-K ranking on contextual multi-armed bandit algorithms via deep learning in order to handle non-linear high-dimensional complex contexts and rewards. The proposed method was evaluated in two experimental scenarios and validated the method with appropriate assumptions. In general, we demonstrated the top-K ranking via deep contextual bandit method learned to select the correct objects over time. This method aims to enable systematic selection of information objects in ranked order, learning from the historical responses, and optimizing the potential use of the information while adapting to the dynamic conditions of the environments.

Even though deep learning can provide flexibility to overcome non-linearity, it can become computationally resource intensive and even degrade performance in some cases. Therefore, deep learning must be considered when such an implementation is practical and valid under careful construction of layers and tuning parameters.

For the future work, we plan to investigate how the $\epsilon-greedy$ and Thompson-sampling, as well as other recently developed bandit algorithms, would perform on a linear model or neural-linear compared to those on DNN. The DNN structure and the hyper-parameters for optimization tuning will be also investigated. Further, it is of interest to investigate the robustness of the models under high-level noise at varying degrees. There has been recent works on diverse ranking in information retrieval research area to minimize  redundant or similar content selection \cite{Radlinski2008LearningDR}. This aspect of diversifying the objects within the top-K ranking is also an important consideration.

\section*{ACKNOWLEDGMENT}
We thank the anonymous referees for their helpful suggestions.


%%%%%%%%%%%%%%%%%%%%%%%%%%%%%%%%%%%%%%%%%%%%%%%%%%%%%%%%%%%%%%%%%%%%%%%%%%%%%%%%

\bibliographystyle{ieeetr}
\bibliography{reference}

\end{document}